%% file: SubmissionID_6096.tex
\documentclass[10pt,twocolumn,letterpaper]{article}

\usepackage{cvpr}
\usepackage{times}
\usepackage{epsfig}
\usepackage{graphicx}
\usepackage{amsmath}
\usepackage{amssymb}

\usepackage[pagebackref=true,breaklinks=true,letterpaper=true,colorlinks,bookmarks=false]{hyperref}
\cvprfinalcopy 

\def\cvprPaperID{6096} 

\usepackage[utf8]{inputenc} 
\usepackage[T1]{fontenc}    
\usepackage{hyperref}       
\usepackage{url}            
\usepackage{booktabs}       
\usepackage{amsfonts}       
\usepackage{amsmath}
\usepackage{amssymb}
\usepackage{nicefrac}       
\usepackage{microtype}      
\usepackage{adjustbox}
\usepackage{booktabs}
\usepackage{multirow}
\usepackage{adjustbox}
\usepackage{xcolor}
\usepackage{graphics,psfrag}
\usepackage{diagbox} 
\usepackage{subfigure}
\usepackage{bbding}
\usepackage{amssymb}

\makeatletter
\ifcvprfinal\fi

\begin{document}
 \pagenumbering{gobble} 
\title{Video Super-resolution with Temporal Group Attention}

\def\@fnsymbol#1{\ensuremath{\ifcase#1\or \dagger\or* \or \ddagger\or
		\mathsection\or \mathparagraph\or \|\or **\or \dagger\dagger
		\or \ddagger\ddagger \else\@ctrerr\fi}}

\author{%
	Takashi Isobe$^{1,2}$\thanks{The work was done in Noah's Ark Lab, Huawei Technologies.}, Songjiang Li$^2$, Xu Jia$^2$$^*$, Shanxin Yuan$^2$, Gregory Slabaugh$^2$, \\ Chunjing Xu$^2$, Ya-Li Li$^1$, Shengjin Wang$^1$\thanks{Corresponding author}, Qi Tian$^2$\\
	{$^1$Department of Electronic Engineering, Tsinghua University}\\
	{$^2$Noah's Ark Lab, Huawei Technologies}\\
	{\texttt{\small{jbj18@mails.tsinghua.edu.cn}} \hspace{0.5cm}}
	{\texttt{\small{\{liyali13, wgsgj\}@tsinghua.edu.cn}}\hspace{0.5cm}} \\
	{\texttt{\small{\{x.jia, songjiang.li, shanxin.yuan, gregory.slabaugh, tian.qi1\}@huawei.com}}}
}

\maketitle

\begin{abstract}
Video super-resolution, which aims at producing a high-resolution video from its corresponding low-resolution version, has recently drawn increasing attention. In this work, we propose a novel method that can effectively incorporate temporal information in a hierarchical way. The input sequence is divided into several groups, with each one corresponding to a kind of frame rate. These groups provide complementary information to recover missing details in the reference frame, which is further integrated with an attention module and a deep intra-group fusion module. In addition, a fast spatial alignment is proposed to handle videos with large motion. Extensive results demonstrate the capability of the proposed model in handling videos with various motion. It achieves favorable performance against state-of-the-art methods on several benchmark datasets. Code is available at \url{https://github.com/junpan19/VSR_TGA}.
\end{abstract}
\input{1-introduction}

\input{2-related}
\input{3-method}
\input{4-experiment}
\input{5-conclusion}

{\small
\bibliographystyle{ieee_fullname}
\bibliography{res}
}

\end{document}

%% file: 1-introduction.tex
\section{Introduction}
\label{intro}

Super-resolution aims at producing high-resolution (HR) images from the corresponding low-resolution (LR) ones by filling in missing details. 
For single image super-resolution, an HR image is estimated by exploring natural image priors and self-similarity within the image. 
For video super-resolution, both spatial information across positions and temporal information across frames can be used to enhance details for an LR frame. 
Recently the task of video super-resolution has drawn much attention in both the research and industrial communities. For example, video super-resolution is required when videos recorded for surveillance are zoomed in to recognize a person's identity or a car's license, or when videos are projected to a high definition display device for visually pleasant watching.

Most video super-resolution methods~\cite{kappeler2016video,caballero2017real,tao2017detail,xue2019video,liu2017robust} adopt the following pipeline: motion estimation, motion compensation, fusion and upsampling. 
They estimate optical flow between a reference frame and other frames in either an offline or online manner, and then align all other frames to the reference with backward warping. However, this is not optimal for video SR. Methods with explicit motion compensation rely heavily on the accuracy of motion estimation. Inaccurate motion estimation and alignment, especially when there is occlusion or complex motion, results in distortion and errors, deteriorating the final super-resolution performance. Besides, per-pixel motion estimation such as optical flow often suffers a heavy computational load.
\begin{figure}[t]
	\centering
	\includegraphics[width=1\columnwidth]{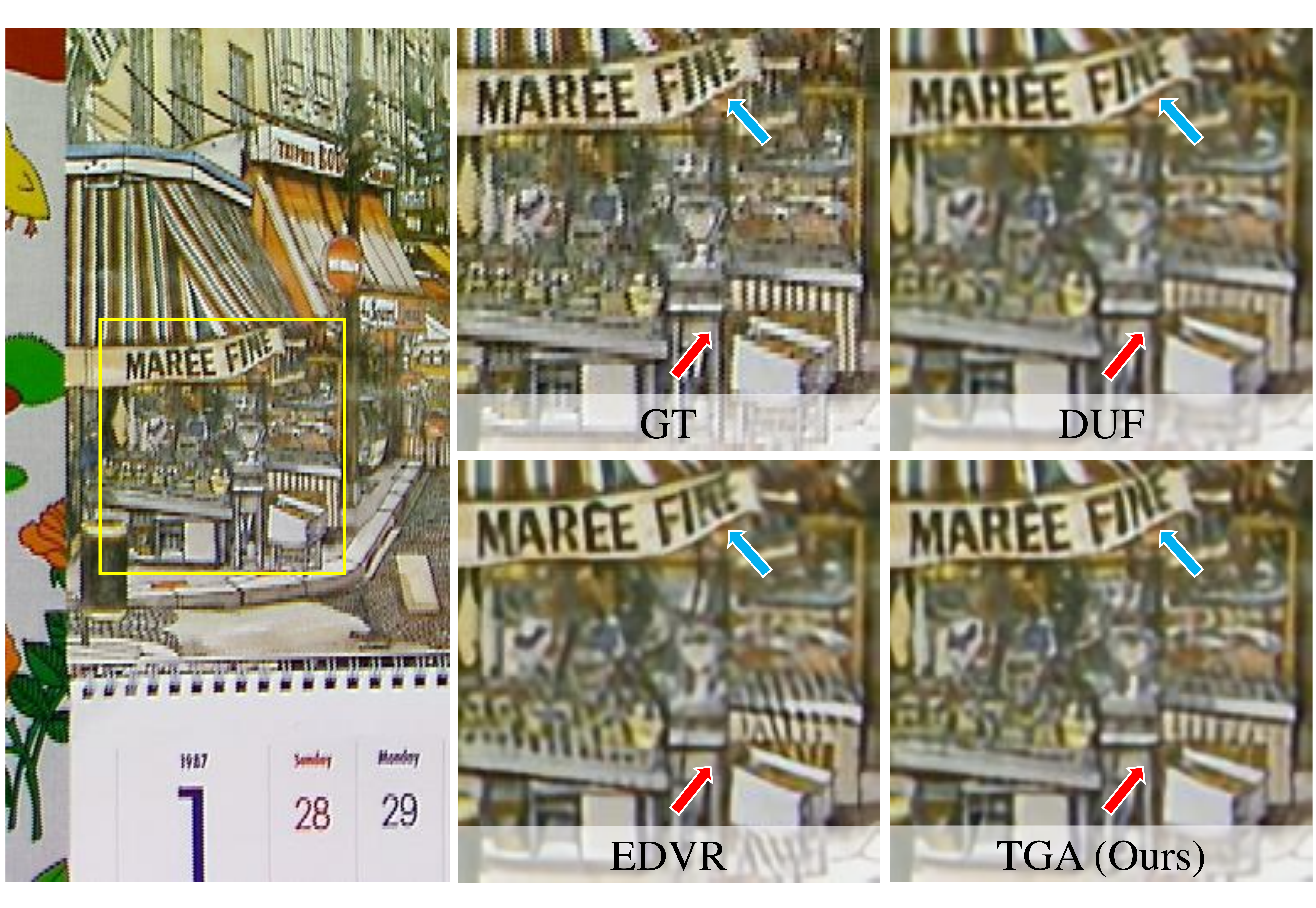}
	\caption{VSR results for the \textit{Calender} clip  in Vid4~\cite{caballero2017real}. Our method produces result with more details ({\color{cyan}cyan} arrow), and fewer artifacts ({\color{red}red} arrow) than DUF~\cite{jo2018deep} and the recent proprosed EDVR~\cite{wang2019edvr}.
	} 	\vspace{-5mm}
	\label{inter-group}
\end{figure} 
Recently Jo~\emph{et al.}~\cite{jo2018deep} proposed the DUF method which implicitly utilizes motion information among LR frames to recover HR frames by means of dynamic upsampling filters. It is less influenced by the accuracy of motion estimation but its performance is limited by the size of the dynamic upsampling filters. In addition, the temporal information integration process from other frames to the reference frame is conducted without explicitly taking the reference frame into consideration. This leads to ineffective information integration for border frames in an input sequence.

In this work, we propose a novel deep neural network which hierarchically utilizes motion information in an implicit manner and is able to make full use of complementary information across frames to recover missing details for the reference frame. Instead of aligning all other frames to the reference frame with optical flow or applying 3D convolution to the whole sequence,we propose to divide a sequence into several groups and conduct information integration in a hierarchical way, that is, first integrating information in each group and then integrate information across groups. The proposed grouping method produces groups of subsequences with different frame rates, which provide different kinds of complementary information for the reference frame. Such different complementary information is modeled with an attention module and the groups are deeply fused with a 3D dense block and a 2D dense block to generate a high-resolution version of the reference frame. Overall, the proposed method follows a hierarchical manner. It is able to handle various kinds of motion and adaptively borrow information from groups of different frame rates. For example, if an object is occluded in one frame, the model would pay more attention to frames in which the object is not occluded. 

However, the capability of the proposed method is still limited in dealing with video sequences of large motion since the receptive field is finite. To address this issue, a fast homography based method is proposed for rough motion compensation among frames. The resulting warped frames are not perfectly aligned but they suffer less distortion artifacts compared to existing optical flow based methods. Appearance difference among frames is indeed reduced such that the proposed neural network model can focus on object motion and produce better super-resolution result.

The proposed method is evaluated on several video super-resolution benchmarks and achieves state-of-the-art performance. We conduct further analysis to demonstrate its effectiveness. 

To sum up, we make the following contributions: 
\begin{itemize}
	\item We propose a novel neural network which efficiently fuses spatio-temporal information through frame-rate-aware groups in a hierarchical manner.
	\item We introduce a fast spatial alignment method to handle videos with large motion.
	\item The proposed method achieves state-of-the-art performance on two popular VSR benchmarks.
\end{itemize}

%% file: 2-related.tex
\section{Related Work}
\label{related}

\subsection{Single Image Super Resolution}
Single image super-resolution (SISR) has benefited greatly from progress in deep learning. Dong~\cite{dong2014learning} first proposed to use a three-layer CNN for SISR and showed impressive potential in super-resolving LR images. New architectures have been designed since then, including a very deep CNN with residual connections~\cite{kim2016accurate}, a recursive architecture with skip-connections~\cite{kim2016deeply}, a architecture with a sub-pixel layer and multi-channel output to directly work on LR images as input~\cite{shi2016real}. More recent networks, including EDSR~\cite{lim2017enhanced}, RDN~\cite{zhang2018residual}, DBPN~\cite{haris2018deep}, RCAN~\cite{zhang2018image}, outperformed previous works by a large margin when trained on the novel large dataset DIV2K~\cite{timofte2017ntire}. More discussions can be found in the recent survey \cite{yang2019deep}.

\subsection{Video Super Resolution}
Video super resolution relies heavily on temporal alignment, either explicitly or implicitly, to make use of complementary information from neighboring low-resolution frames. VESCPN \cite{caballero2017real} is the first end-to-end video SR method that jointly trains optical flow estimation and spatial-temporal networks. SPMC \cite{tao2017detail} proposed a new sub-pixel motion compensation layer for inter-frame motion alignment, and achieved motion compensation and upsampling simultaneously. \cite{xue2019video} proposed to jointly train the motion analysis and video super resolution in an end-to-end manner through a proposed task-oriented flow. \cite{haris2019recurrent} proposed to use a recurrent encoder-decoder module to exploit spatial and temporal information, where explicit inter-frame motion were estimated. Methods using implicit temporal alignment showed superior performance on several benchmarks. \cite{kim20183dsrnet} exploited the 3DCNN's spatial-temporal feature representation capability to avoid motion alignment, and stacked several 3D convolutional layers for video SR. \cite{jo2018deep} proposed to use 3D convolutional layers to compute dynamic filters~\cite{jia2016dynamic} for implicit motion compensation and upsampling. Instead of image level motion alignment, TDAN~\cite{tian2018tdan} and EDVR~\cite{wang2019edvr} worked in the feature level motion alignment. TDAN \cite{tian2018tdan} proposed a temporal deformable alignment module to align features of different frames for better performance. EDVR \cite{wang2019edvr} extended TDAN in two aspects by 1) using deformable alignment in a coarse-to-fine manner and 2) proposing a new temporal and spatial attention attention fusion module, instead of naively concatenating the aligned LR frames as TDAN does. 

The work most related with ours is~\cite{liu2017robust}, which also re-organized the input frames to several groups. However, in~\cite{liu2017robust}, groups are composed of different number of input frames. In addition, that method generates an super-resolution result for each group and computes an attention map to combine these super-resolution results, which takes much computation and is not very effective. Our method divides input frames into several groups based on frame rate and effectively integrates temporal information in a hierarchical way.
\begin{figure*}[t]
	\centering
	\includegraphics[width=1\textwidth]{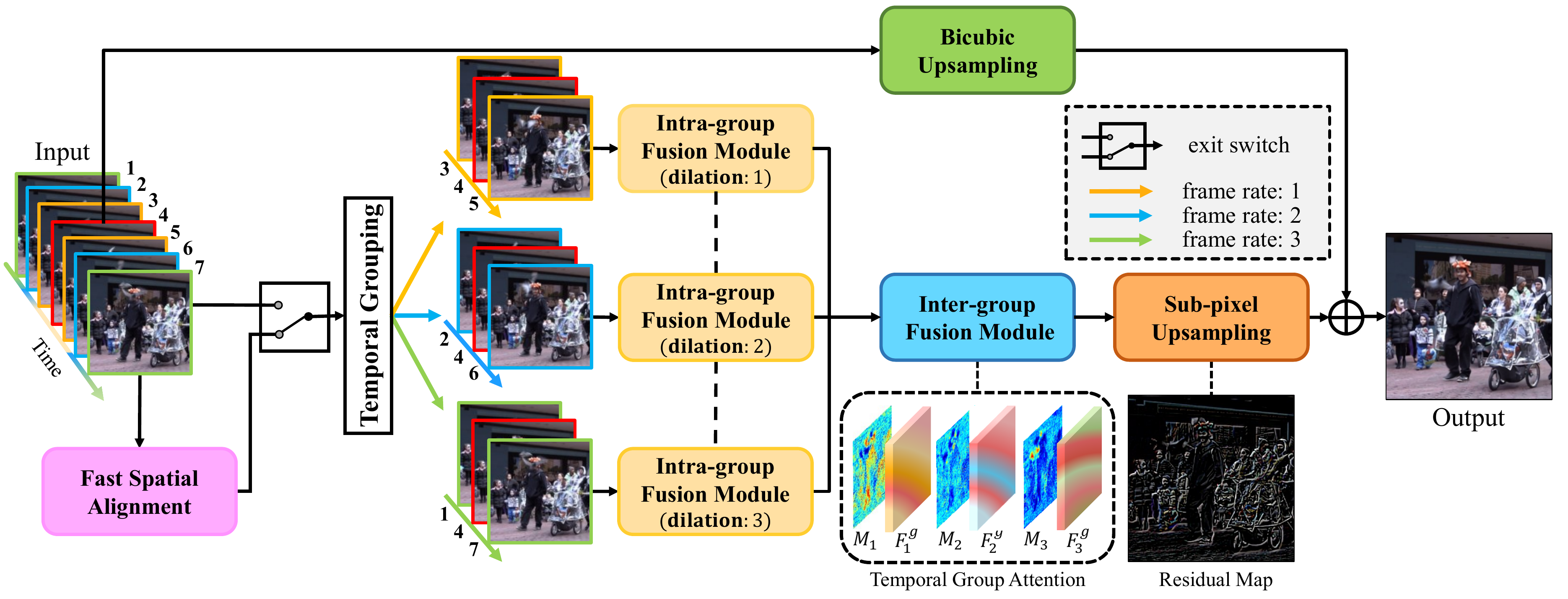}
	\caption{The proposed method with temporal group attention. }
	\vspace{-2mm}
	\label{pipeline}
\end{figure*}

%% file: 3-method.tex
\section{Methodology}
\label{method}

\subsection{Overview}

Given a consecutive low-resolution video frame sequence consisting of one reference frame $I_t^L$ and $2N$ neighboring frames \{$I_{t-N}^L:I_{t-1}^L,I_{t+1}^L:I_{t+N}^L$\}, the goal of VSR is to reconstruct a high-resolution version of reference frame $\hat{I_{t}}$ by fully utilizing the spatio-temporal information across the sequence.
The overall pipeline of the proposed method is shown in Fig.~\ref{pipeline}. It's a generic framework suitable for processing sequences of different input lengths. Take seven frames $\{I_1^L, I_2^L, ..., I_7^L\}$ for example, we denote the middle frame $I_4^L$ as the reference frame, and the other frames as neighboring ones. 
The seven input frames are divided into three groups based on decoupled motion, with each one representing a certain kind of frame rate. An intra-group fusion module with shared weights is proposed to extract and fuse spatio-temporal information within each group. Information across groups is further integrated through an attention-based inter-group fusion module. Finally, the output high-resolution frame $\hat{I_4}$ is generated by adding the network produced residual map and the bicubic upsampling of the input reference frame. 
Additionally, a fast spatial alignmemt module is proposed to further help deal with video sequences of large motion. 


\subsection{Temporal Group Attention} 
The crucial problem with implicit motion compensation lies on the inefficient fusion of temporal fusion in neighboring frames. In~\cite{jo2018deep}, input frames are stacked along the temporal axis and 3D convolutions are directly applied to the stacked frames. Such distant neighboring frames are not explicitly guided by the reference frame, resulting in insufficient information fusion, and this impedes the reference frame from borrowing information from distant frames. To address this issue, we propose to split neighboring $2N$ frames into N groups based on their temporal distances from the reference frame. Later, spatial-temporal information is extracted and fused in a hierarchical manner: an intra-group fusion module integrates information within each group, followed by an inter-group fusion module which effectively handles group-wise features.

\textbf{Temporal Grouping.} In contrast to the previous work, the neighboring $2N$ frames are split to $N$ groups based on the temporal distance to the reference frame. 
The original sequence is reordered as $\{G_1,...,G_n\}$, $n\in[1:N]$, where $G_n=\{I^L_{t-n}, I_t^L,I^L_{t+n}\}$ is a subsequence consisting of a former frame $I_{t-n}^L$, the reference frame $I_{t}^L$ and a latter frame $I_{t+n}^L$. Notice that the reference frame appears in each group. It is noteworthy that our method can be easily generalized to arbitrary frames as input.
The grouping allows explicit and efficient integration of neighboring frames with different temporal distance for two reasons: 
1) The contributions of neighboring frames in different temporal distances are not equal, especially for frames with large deformation, occlusion and motion blur. 
When a region in one group is (for example by occlusion), the missing information can be recovered by other groups. That is, information of different groups complements each other. 2) 
The reference frame in each group guides the model to extract beneficial information from neighboring frames, allowing efficient information extraction and fusion.

\textbf {Intra-group Fusion.} For each group, an intra-group fusion module is deployed for feature extraction and fusion within each group. 
The module consists of three parts. The first part contains three units as the spatial features extractor, where each unit is composed of a $3\times 3$ convolutional layer followed by a batch normalization (BN) \cite{normalization2015accelerating} and a ReLU \cite{glorot2011deep}. All convolutional layers are equipped with dilation rate to model the motion level associated with a group. The dilation rate is determined according to the frame rate in each group with the assumption that distant group has large motion and near group has small motion. Subsequently, for the second part, an additional 3D convolutional layer with $3\times3\times3$ kernel is used to perform spatio-temporal feature fusion. Finally, group-wise features $F_n^g$ are produced by applying eighteen 2D units in the 2D dense block to deeply integrate information within each group.


The weights of the intra-group fusion module are shared for each group for efficiency. The effectiveness of the proposed temporal grouping are presented in Sec.4.3.

\textbf{Inter-group Fusion with Temporal Attention.} To better integrate features from different groups, a temporal attention module is introduced. Temporal attention has been widely used in video-related tasks~\cite{song2017end,zanfir2016spatio,zang2018attention,yan2019stat}. In this work, we show that temporal attention also benefits the task of VSR by enabling the model to pay different attention across time. In the previous section, a frame sequence is categorized into groups according to different frame rates. These groups contain complementary information. Usually, a group with slow frame rate is more informative because the neighboring frames are more similar to the reference one. Simultaneously, groups with fast frame rate may also capture information about some fine details which are missing in the nearby frames. Hence, temporal attention works as a guidance to efficiently integrate features from different temporal interval groups.
\begin{figure}[thbp]
	\label{2}
	\centering
	\includegraphics[width=1\columnwidth]{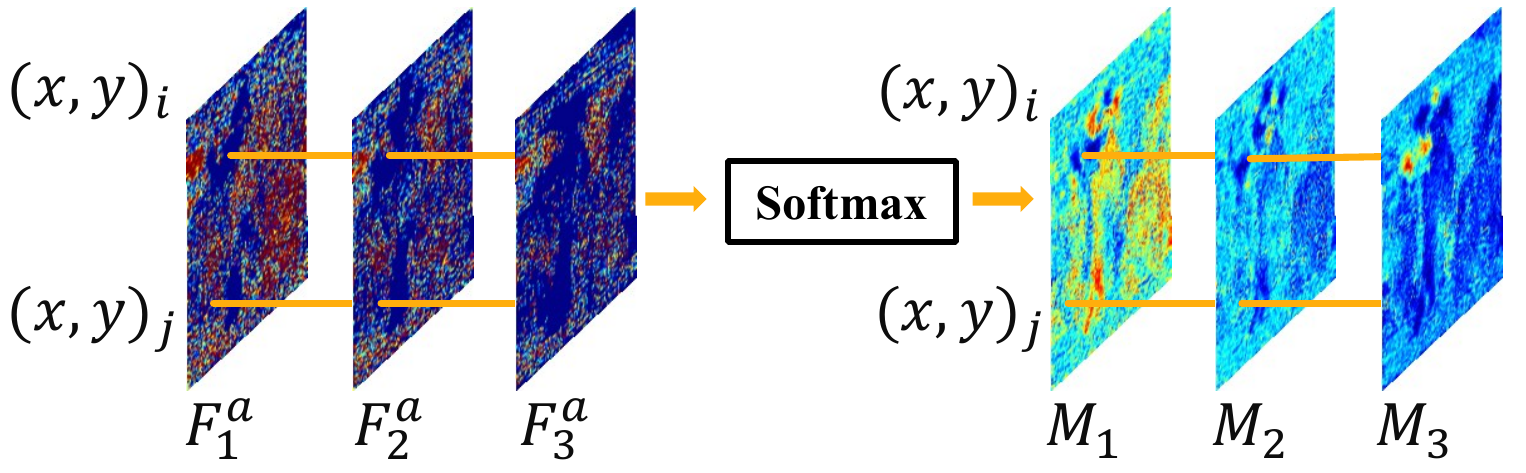}
	\caption{Computation of group attention maps. $F_n^a$ corresponds to group-wise features while $M_n$ is the attention mask.}
	\vspace{-1mm}
	\label{softmax}
\end{figure}

For each group, a one-channel feature map $F^a_n$ is computed after applying a $3\times3$ convolutional layer on top of the corresponding feature maps $F^g_n$. They are further concatenated and a softmax function along temporal axis is applied to each position across channels to compute attention maps, as shown in Fig.~\ref{softmax}. Each group's intermediate map is concatenated and the attention maps $M(x,y)$ are computed by applying softmax along temporal axis, as shown in Fig.~\ref{softmax}. 
\begin{equation}
\label{4}
M_n(x,y)_j=\dfrac{e^{F^{a}_{n}(x,y)_j}}{\sum_{i=1}^N e^{F_i^a(x,y)_j}}
\end{equation}
Attention weighted feature for each group $\widetilde{F}^g_n$ is calculated as:
\begin{equation}
	\label{5}
	\widetilde{F}^g_n = M_n \odot {F_n^g}, n\in[1:N]
\end{equation}
where $M_n(x,y)_j$ represents the weight of the temporal group attention mask at location $(x,y)_j$. ${F_n^g}$ represents the group-wise features produced by intra-group fusion module. `$\odot$' denotes element-wise multiplication.

\begin{figure}[t]
	\centering
	\includegraphics[width=1\columnwidth]{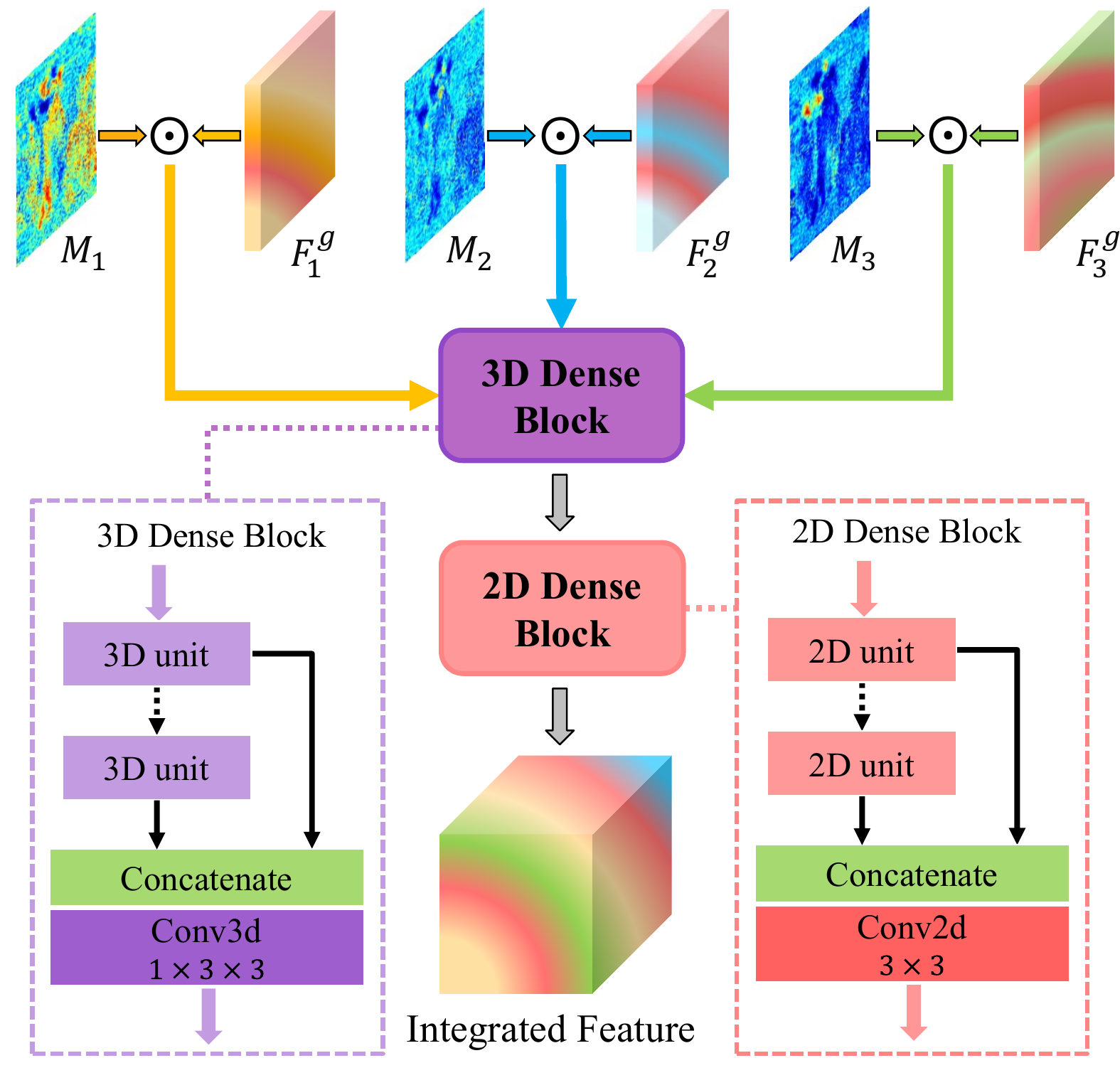}
	\caption{Structure of the inter-group fusion module. 
	}
	\vspace{-3mm}
	\label{inter-group}
\end{figure}

The goal of the inter-group fusion module is to aggregate information across different temporal groups and produce a high-resolution residual map. 
In order to make full use of attention weighted feature over temporal groups, we first aggregate those features by concatenating them along the temporal axis and feed it into a 3D dense block. Then a 2D dense block is on top for further fusion, as shown in Fig.~\ref{inter-group}. 3D unit has the same structure as 2D unit which is used in intra-group fusion module. A convolution layer with $1\times3\times3$ kernel is inserted in the end of the 3D dense block to reduce channels. The design of 2D and 3D dense blocks are inspired by RDN~\cite{zhang2018residual} and DUF~\cite{jo2018deep}, which is modified in an efficient way to our pipeline.

Finally, similar to several single image super-resolution methods, sufficiently aggregated features are upsampled with a depth-to-space operation~\cite{shi2016real} to produce high-resolution residual map $R_t$. The high-resolution reconstruction $\hat{I}_t$ is computed as the sum of the residual map $R_t$ and a bicubic upsampled reference image $I^{\uparrow}_t$.

\subsection{Fast Spatial Alignment}
\begin{figure*}[t]
	\label{1}
	\centering
	\includegraphics[width=\textwidth]{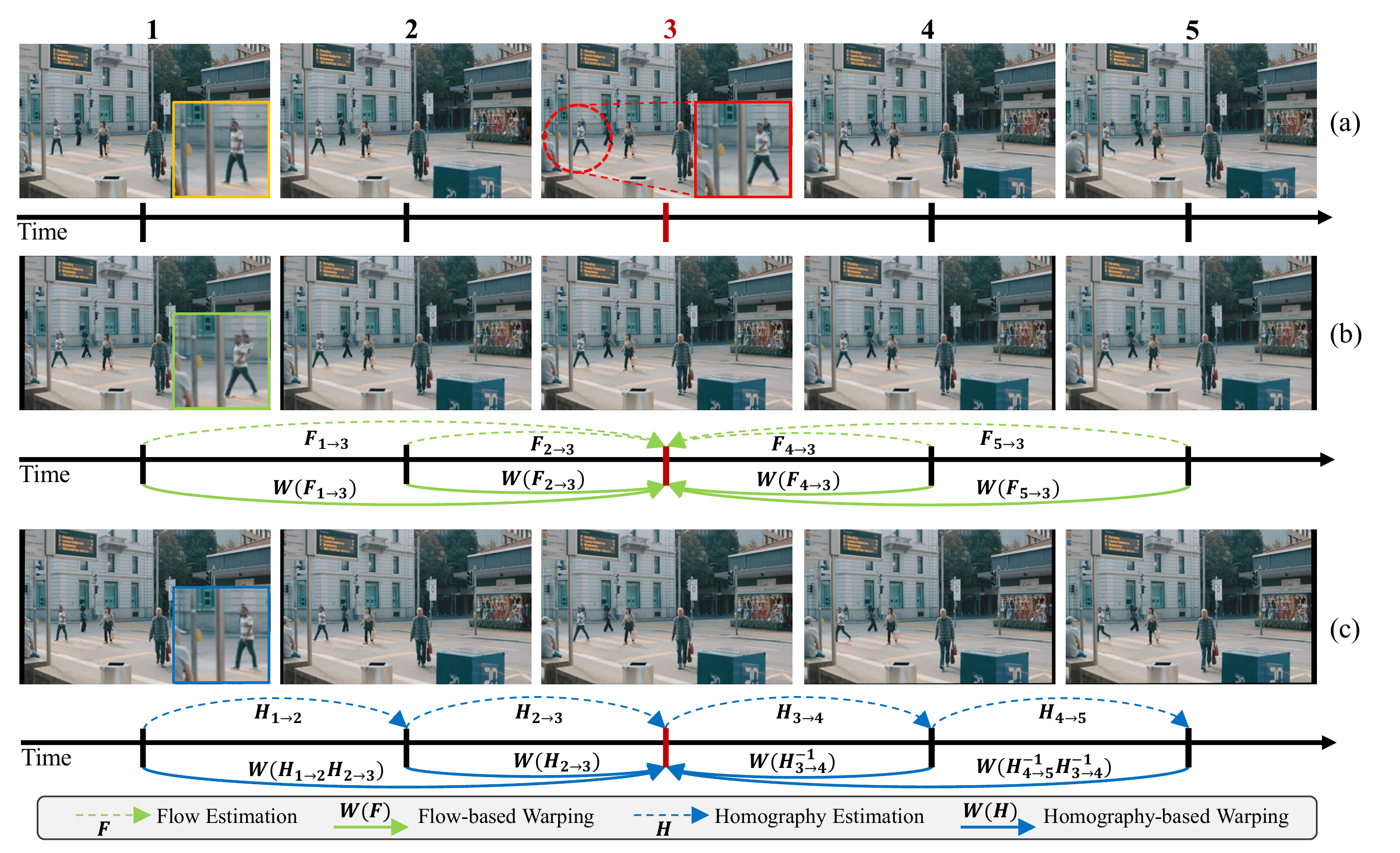}
	\caption{Fast spatial alignment compared with optical flow. (a) Original 5 consecutive frames, of which \textbf{\color{red}frame 3} is the reference frame. (b) Alignment with optical flow. The flow for each neighboring frame is estimated independently. (c) The proposed alignment only estimates basic homographies for consecutive frames. The frame-level alignment suppresses pixel-level distortion. Zoom in for better visualization.}
	\vspace{-2mm}
	\label{flow_homo}
\end{figure*}

Although the proposed model is able to effectively use temporal information across frames, it has difficulty in dealing with videos with large motion. To improve the performance of the proposed model in case of large motion, we further propose a fast spatial alignment module. Different from previous methods~\cite{ma2015handling,caballero2017real,xue2019video} which either use offline optical flow or an integrated optical flow network for motion estimation and compensation, we estimate homography between every two consecutive frames and warp neighboring frames to the reference frame, which can be shown in Fig.~\ref{flow_homo}. 
Interest points could be detected by feature detectors such as SIFT~\cite{lowe2004distinctive} or ORB~\cite{rublee2011orb}, and point correspondences are computed to estimate homography.
Homography from frame $A$ and $C$ can be computed as a product of the homography from $A$ to $B$ and the one from $B$ to $C$:
\begin{equation}
H_{A\to C} = H_{A\to B}\cdot H_{B\to C}
\end{equation}
For a homography, the inverse transform can be represented by the inverse of the matrix:
\begin{equation}
H_{B\to A} = H_{A\to B}^{-1}
\end{equation}
Since optical flow is computed for each pixel, imperfect optical flow estimation would introduce much unexpected pixel-level distortion into warping, destroying structure in original images. In addition, most optical-flow-based methods~\cite{liao2015video,caballero2017real,tao2017detail,xue2019video} estimate optical flow between each neighboring frame and the reference frame independently, which would bring a lot of redundant computation  when super-resolving a long sequence. In our method, since homography transformation is a global, it keeps the structure better and introduces little artifact. In addition, the associative composition nature of homography allows to decompose a homography between two frames into a product of homographies between every two consecutive ones in that interval, which avoids redundant computation and speeds up pre-alignment. Note that the pre-alignment here does not need to be perfect. As long as it does not introduce much pixel-level distortion, the proposed VSR network can give good performance. We also introduce exit mechanism for pre-alignment for robustness. That is, in case that few interest points are detected or there is much difference between a frame and the result after applying $H$ and $H^{-1}$, the frames are kept as they are without any pre-alignment. In other words, a conservative strategy is adopt in pre-alignment procedure.

%


%% file: 4-experiment.tex
\section{Experiments}
\label{experiment}
\begin{table*}[t]
	\centering
	\scalebox{0.9}{	
		\begin{tabular}{lccccccc}
			\toprule
			Method    &\# Frames & Calendar (Y) & City (Y) & Foliage (Y)  & Walk (Y) &Average (Y)  & Average (RGB)  
			\\
			\midrule    	
			Bicubic & 1 &18.83/0.4936  &23.84/0.5234	&21.52/0.4438  &23.01/0.7096  						 	    & 21.80/0.5426   & 20.37/0.5106	
			\\	
			SPMC $^\dagger$~\cite{tao2017detail}  & 3 & -  &- &-  &- & 25.52/0.76~~~~	&-
			\\
			Liu$^\dagger$~\cite{liu2017robust} & 5  &21.61/~~~~~-~~~~~ &26.29/~~~~~-~~~~~ &24.99/~~~~~-~~~~~ &28.06/~~~~~-~~~~~ &25.23/~~~~~-~~~~~  & -
			\\
			TOFlow~\cite{xue2019video} & 7 & 22.29/0.7273  & 26.79/0.7446 & 25.31/0.7118 & 29.02/0.8799 & 25.85/0.7659 &24.39/0.7438	
			\\
			FRVSR~$^\dagger$\cite{sajjadi2018frame} &recurrent  & - &- &-  &- & 26.69/0.822~~ &-	
			\\
			DUF-52L~\cite{jo2018deep} & 7 &24.17/0.8161 &28.05/0.8235 &26.42/0.7758 & 30.91/ \textbf{\color{blue}0.9165}	&27.38/0.8329 &\textbf{\color{blue} 25.91}/\textbf{\color{blue}0.8166}
			\\
			RBPN~\cite{haris2019recurrent} & 7 &24.02/0.8088 &27.83/0.8045 &26.21/0.7579 &30.62/0.9111 &27.17/0.8205  &25.65/0.7997 
			\\
			EDVR-L$^\dagger$~\cite{wang2019edvr} & 7 & 24.05/0.8147 &28.00/0.8122	 &26.34/0.7635  &\textbf{\color{red}31.02}/0.9152  & 27.35/0.8264  &25.83/0.8077   	
			\\		
			PFNL$^\dagger$~\cite{yi2019progressive} &7 &\textbf{\color{blue} 24.37}/\textbf{\color{blue}0.8246} &\textbf{\color{blue} 28.09}/\textbf{\color{blue}0.8385} &\textbf{\color{blue}26.51}/\textbf{\color{blue}0.7768} &30.65/0.9135 &\textbf{\color{blue}27.40}/\textbf{\color{blue}0.8384} &-
			\\
			TGA (Ours)  & 7 &\textbf{\color{red}24.47}/\textbf{\color{red}0.8286} &\textbf{\color{red}28.37}/\textbf{\color{red}0.8419} & \textbf{\color{red}26.59}/\textbf{\color{red}0.7793} &\textbf{\color{blue}30.96}/\textbf{\color{red}0.9181} &\textbf{\color{red}27.59}/\textbf{\color{red}0.8419} &\textbf{\color{red}26.10}/\textbf{\color{red}0.8254}
			\\
			\bottomrule
		\end{tabular}
	}
	\vspace{3mm}
	\caption{Quantitative comparison (PSNR(dB) and SSIM) on \textbf{Vid4} for $4\times$ video super-resolution. {\color{red}Red} text indicates the best and {\color{blue} blue} text indicates the second best performance. Y and RGB indicate the luminance and RGB channels, respectively. `$\dagger$' means the values are taken from original publications or calculated by provided models. Best view in color.}
	\label{vid4_table}
\end{table*}

\begin{table*}[t]
	\centering
	\scalebox{0.92}{	
		\begin{tabular}{lcccccc}
			\toprule
			&Bicubic   &TOFlow~\cite{xue2019video}  &DUF-52L ~\cite{jo2018deep}   &RBPN~\cite{haris2019recurrent}  &EDVR-L$^\dagger$~\cite{wang2019edvr}  &TGA(Ours)
			\\
			\midrule 
			\# Param.  &N/A &1.4M  &5.8M  &12.1M &20.6M &5.8M
			\\
			FLOPs    &N/A  &0.27T &0.20T & 3.08T  &0.30T  &0.07T
			\\   	
			Y Channel  &31.30/0.8687 &34.62/0.9212    &36.87/0.9447 &37.20/0.9458   &\textbf{\color{red} 37.61}/\textbf{\color{blue}0.9489} &\textbf{\color{blue}37.59}/\textbf{\color{red}0.9516}
			\\
			RGB Channels &29.77/0.8490 &32.78/0.9040    &34.96/0.9313 &35.39/0.9340	&\textbf{\color{red} 35.79}/\textbf{\color{blue} 0.9374}  &\textbf{\color{blue}35.57}/\textbf{\color{red} 0.9387}
			\\	
			\bottomrule
		\end{tabular}
	}
	\vspace{3mm}
	\caption{Quantitative comparison (PSNR(dB) and SSIM) on \textbf{Vimeo-90K-T} for $4\times$ video super-resolution. {\color{red}Red} text indicaktes the best result and {\color{blue} blue} text indicates the second best. FLOPs are calculated on an LR image of size 112$\times$64. `$\dagger$' means the values are taken from original publications. Note that the deformation convolution and offline pre-alignment are not included in calculating FLOPs. Best view in color.}
	\vspace{-5mm}
	\label{vimeo_table}
\end{table*}

\begin{figure*}[thbp]
	\centering
	\includegraphics[width=\textwidth]{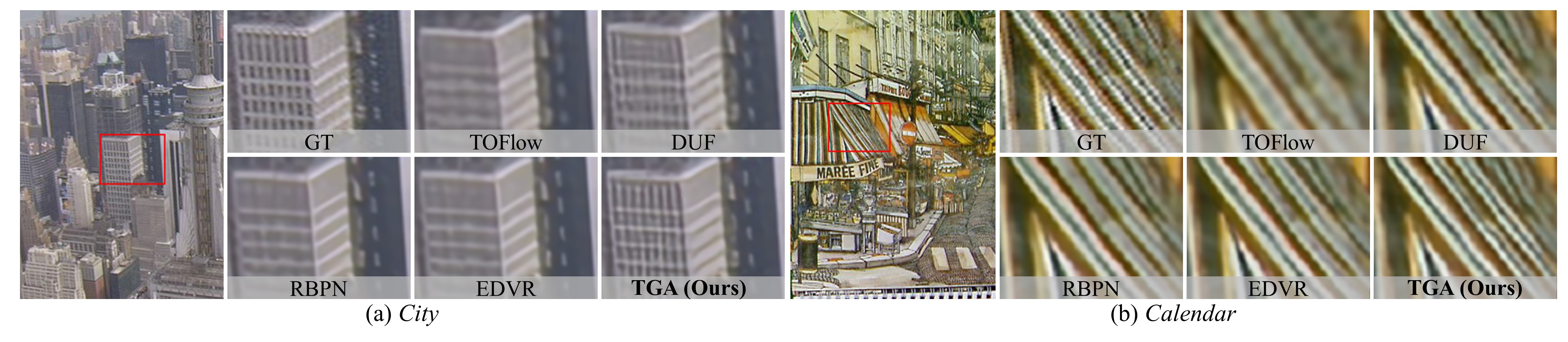}
	\caption{Qualitative comparison on the \textbf{Vid4} for 4$\times$SR. Zoom in for better visualization.}
	\vspace{-2mm}
	\label{vid_figure}
\end{figure*} 

\begin{figure*}[thbp]
	\centering
	\includegraphics[width=\textwidth]{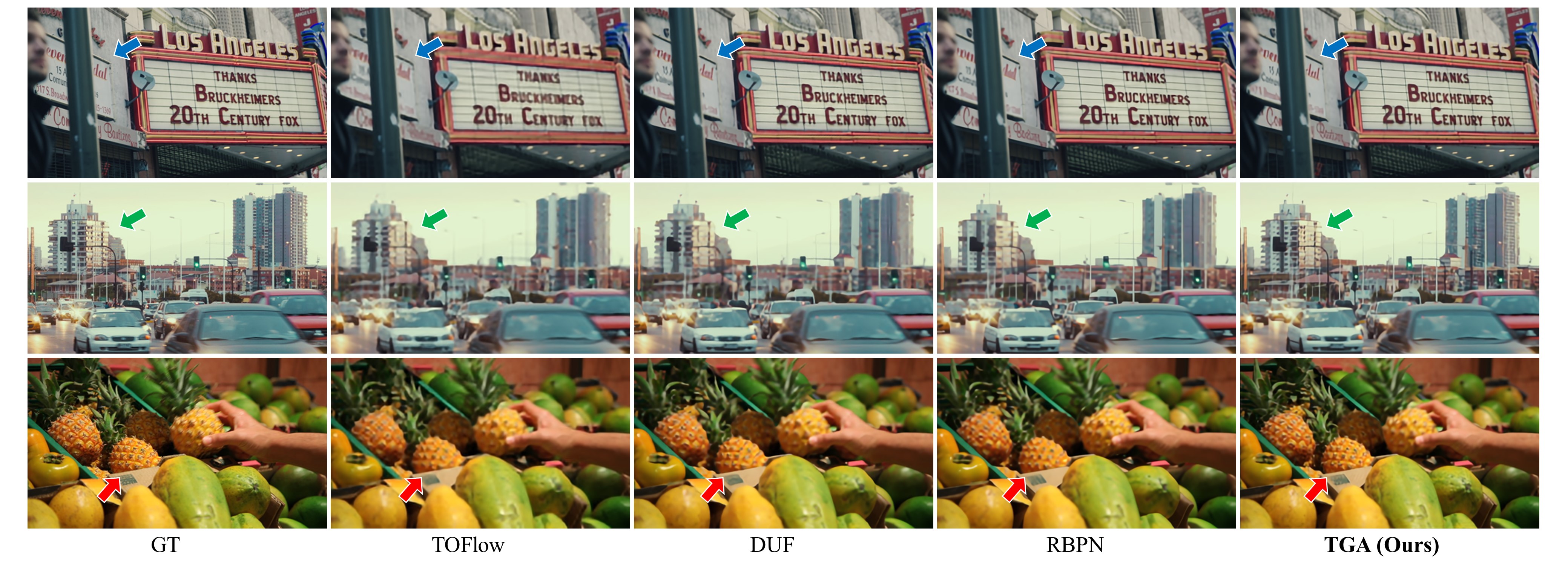}
	\caption{Qualitative comparison on the \textbf{Vimeo-90K-T} for 4$\times$SR. Zoom in for better visualization.}
	\vspace{-3mm}
	\label{vimeo_figure}
\end{figure*} 

To evaluate the proposed method, a series of experiments are conducted and results are compared with existing state-of-the-art methods. Subsequently, a detailed ablation study is conducted to analyze the effectiveness of the proposed temporal grouping, group attention and fast spatial alignment. Results demonstrate the effectiveness and superiority of the proposed method.

\subsection{Implementation Details}
\textbf{Dataset.}
Similar to \cite{haris2019recurrent,xue2019video}, we adopt Vimeo-90k~\cite{xue2019video} as our training set, which is a widely used for the task of video super-resolution. 
We sample regions with spatial resolution 256$\times$256 from high resolution video clips. Similar to \cite{jo2018deep,xue2019video,yi2019progressive} low-resolution patches of $64\times64$ are generated by applying a Gaussian blur with a standard deviation of $\sigma=1.6$ and $4\times$ downsampling. We evaluate the proposed method on two popular benchmarks: Vid4~\cite{liu2013bayesian} and Vimeo-90K-T\cite{xue2019video}. Vid4 consists of four scenes with various motion and occlusion. Vimeo-90K-T contains about 7$k$ high-quality frames and diverse motion types.

\textbf{Implementation details.} 
In the intra-group fusion module, three 2D units are used for spatial features extractor, which is followed by a 3D convolution and eighteen 2D units in the 2D dense block to integrate information within each group.
For the inter-group fusion module, we use four 3D units in the 3D dense block and twenty-one 2D units in the 2D dense block. The channel size is set to 16 for convolutional layers in the 2D and 3D units. Unless specified otherwise, our network takes seven low resolution frames as input. 
The model is supervised by pixel-wise $L1$ loss and optimized with Adam \cite{kingma2014adam} optimizer in which $\beta_1=0.9$ and $\beta_2=0.999$. Weight decay is set to $5\times10^{-4}$ during training. 
The learning rate is initially set to $2\times10^{-3}$ and later down-scaled by a factor of 0.1 every 10 epoches until 30 epochs. 
The size of mini-batch is set to 64. The training data is augmented by flipping and rotating with a probability of 0.5.
All experiments are conducted on a server with Python 3.6.4, PyTorch 1.1 and Nvidia Tesla V100 GPUs.
\subsection{Comparison with State-of-the-arts}
%

\begin{figure}[t]
	\centering
	\includegraphics[width=1\columnwidth]{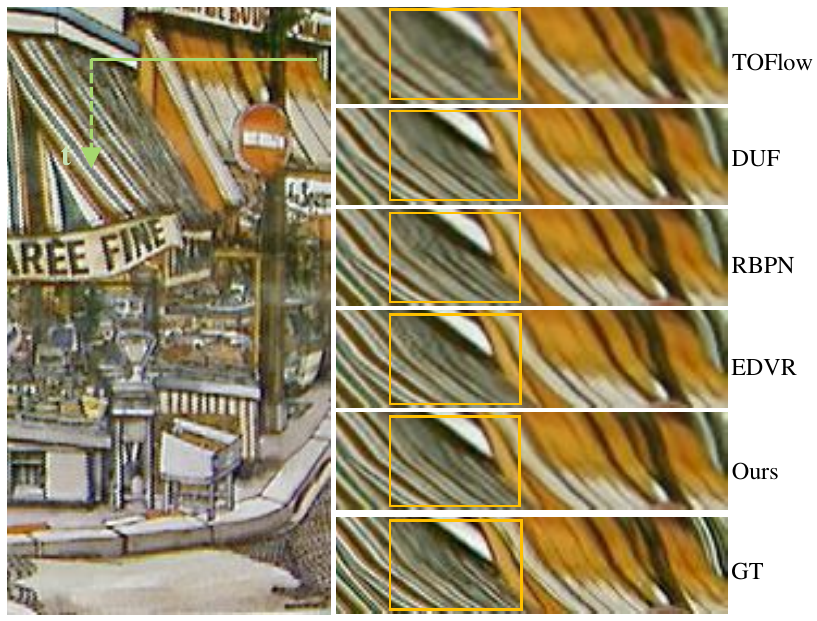}
	\caption{Visualization of temporal consistency for \textit{calendar} sequence. Temporal profile is produced by recording a single pixel line ({\color{green}green} line) spanning time and stacked vertically.
	}
	\vspace{-5mm}
	\label{temporal-profile}
\end{figure} 

We compare the proposed method with six state-of-the-art VSR approaches, including TOFlow~\cite{xue2019video}, SPMC~\cite{tao2017detail}, Liu~\cite{liu2017robust}, DUF~\cite{jo2018deep}, RBPN~\cite{haris2019recurrent}, EDVR~\cite{wang2019edvr} and PFNL~\cite{yi2019progressive}. Both TOFlow and SPMC apply 
explicit pixel-level motion compensation with optical flow estimation, while RBPN uses pre-computed optical flow as additional input. DUF, EDVR and PFNL conduct VSR with implicit motion compensation. We carefully implement TOFlow and DUF on our own, and rebuild RBPN and EDVR based on the publicly available code. We reproduce the performance of most of these methods as reported in the paper except for EDVR. 
Tab.~\ref{vid4_table} and Tab.~\ref{vimeo_table} give quantitative results of state-of-the-art methods on Vid4 and Vimeo-90K-T, which are either reported in the original papers or computed by us. In the evaluation, we take all frames into account except for the DUF method~\cite{jo2018deep} which crop 8 pixels on four borders of each frame since it suffer from severe border artifacts. In addition, we also include the number of parameters and FLOPs for most methods on an LR image of size $112\times64$ in Tab.~\ref{vimeo_table}. 
On Vid4 test set, the proposed method achieves a result of 27.59dB PSNR in the Y channel and 26.10dB PSNR in RGB channel, which outperforms other state-of-the-art methods by a large margin. 
Qualitative result in Fig.~\ref{vid_figure} also validates the superiority of the proposed method. Attributed to the proposed temporal group attention, which is able to make full use of complementary information among frames, our model produces sharper edges and finer detailed texture than other methods.
In addition, we extract temporal profiles in order to evaluate the performance on temporal consistency in Fig.\ref{temporal-profile}. A temporal profile is produced by taking the same horizontal row of pixels from consecutive frames and stacking them vertically. The temporal profiles show that the proposed method gives temporally consistent results, which suffer less flickering artifacts than other approaches.

Vimeo-90K-T is a large and challenging dataset covering scenes with large motion and complicated illumination changes. The proposed method is compared with several methods including TOFlow, DUF, RBPN and EDVR. 
As shown in Tab.~\ref{vimeo_table} and Fig.~\ref{vimeo_figure}, the proposed method also achieves very good performance on this challenging dataset. It outperforms most state-of-the-art methods such as TOFlow, DUF and RBPN by a large margin both in PSNR and SSIM. The only exception is EDVR-L whose model size and computation is about four times larger than our method. In spite of this, our method is still rather comparable in PSNR and a little better in SSIM.

\subsection{Ablation Study}
In this section, we conduct several ablation study on the proposed temporal group attention and fast spatial alignment to further demonstrate the effectiveness of our method.

\textbf{Temporal Group Attention.} First we experiment with different ways of organizing the input sequence. 
One baseline method is to simply stack input frames along temporal axis and directly feed that to several 3D convolutional layers, similar to DUF~\cite{jo2018deep}. Apart from our grouping method $\{345, 246, 147\}$, we also experiment with other ways of grouping: $\{123, 345, 567\}$ and $\{345, 142, 647\}$. As shown in Tab.~\ref{group_mannar}, DUF-like input performs worst among these methods. That illustrate that integrating temporal information in a hierarchical manner is a more effective way in integrating information across frames. Both $\{345, 246, 147\}$ and $\{345, 142, 647\}$ are better than $\{123, 345, 567\}$, which implies the advantage of adding the reference frame in each group. Having the reference in the group encourages the model to extract complementary information that is missing in the reference frame. Another 0.05dB improvement of our grouping method $\{345, 246, 147\}$ could be attributed to the effectiveness of motion-based grouping in employing temporal information.
\begin{table}[t]
	
	\centering
	\scalebox{0.65}{	
		\begin{tabular}{lcccc}
			
			\toprule
			Model   & DUF-like &$\{123,345,567\}$ &$\{345,142,647\}$ &$\{345,246,147\}$  
			\\
			TG?   &\XSolidBrush &\Checkmark  &\Checkmark &\Checkmark
			\\
			\midrule
			Vid4          &27.18/0.8258   &27.47/0.8384  &27.54/0.8409   & \textbf{27.59}/\textbf{0.8419}
			\\
			Vimeo-90K-T   &37.06/0.9465 &37.46/0.9487  &37.51/0.9509  &\textbf{37.59}/\textbf{0.9516}
			\\
			\bottomrule
		\end{tabular}
	}
\vspace{1mm}
	\caption{Ablation on: different grouping strategies.}
	\vspace{-5mm}
	\label{group_mannar}
\end{table}

\begin{figure}[t]
 	\centering
 	\includegraphics[width=1\columnwidth]{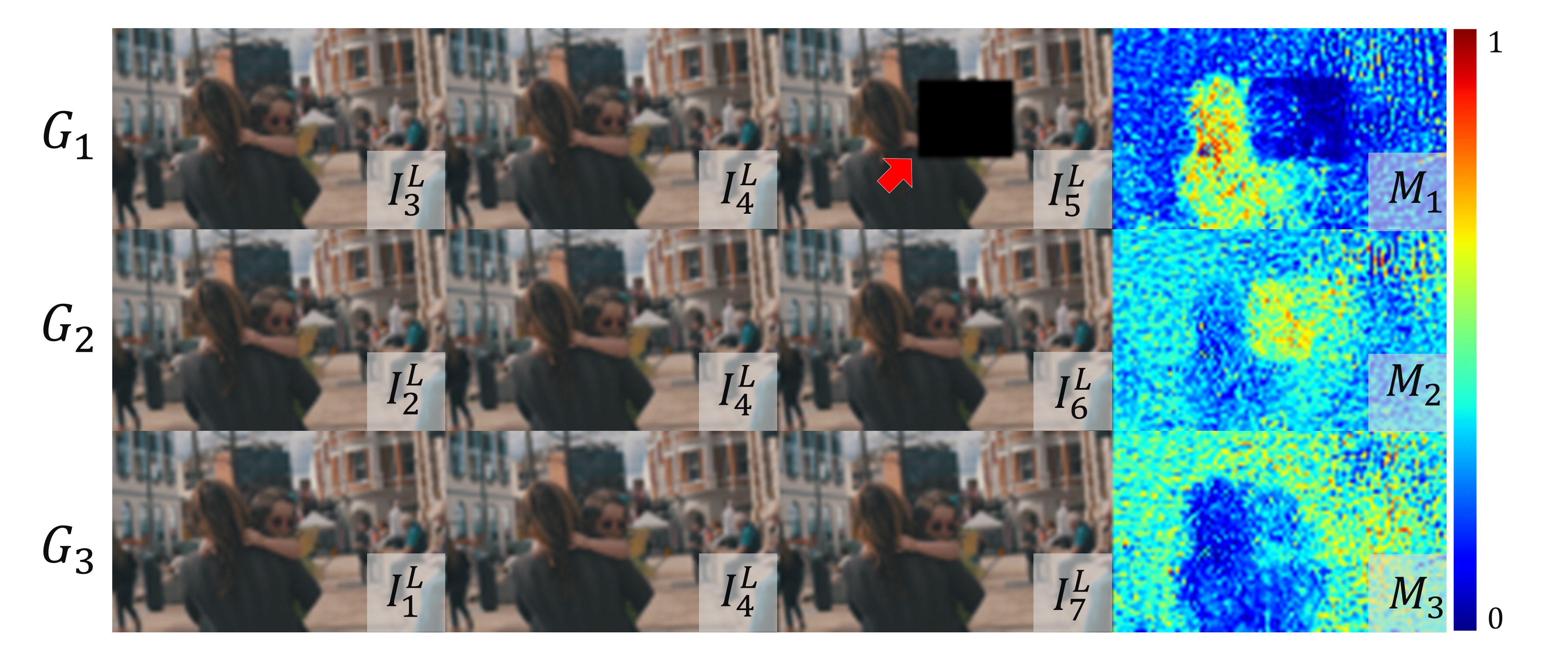}
 	\caption{Visualization of group attention masks under occlusion settings. $G_1, G_2$ and $G_3$ denote three groups. 
 	}
  	\vspace{-5mm}
 	\label{occlusion}
 \end{figure}

In addition, we also evaluate a model which removes the attention module from our whole model. 
As shown in Tab.~\ref{frames}, this model performs a little worse than our full model. 
We also train our full model with a sequence of 5 frames as input. The result in Tab.~\ref{frames} shows that the proposed method can effectively borrow information from additional frames. We notice that the proposed method outperforms DUF even with 2 fewer frames in the input.
In addition, we conduct a toy experiment where a part of a neighboring frame is occluded and visualize the maps of temporal group attention. As shown in Fig.~\ref{occlusion}, the model does attempt to borrow more information from other groups when a group can not provide complementary information to recover the details of that region. 
\begin{table}[th]
	\begin{center}
		\scalebox{0.8}{	
			\begin{tabular}{lccc}
				
				\toprule
				Model   & Model 1 & Model 2 & Model 3
				\\
				\midrule
				\# Frames  & 7   & 5 & 7
				\\
				GA?       &\XSolidBrush &\Checkmark  &\Checkmark
				\\
				\midrule  	
				Vid4       & 27.51/0.8394       &27.39/0.8337   & \textbf{27.59}/\textbf{0.8419}
				\\
				Vimeo-90K-T  & 37.43/0.9506      & 37.34/0.9491   &\textbf{37.59}/\textbf{0.9516}  
				\\
				\bottomrule
			\end{tabular}
		}
	\end{center}
    \vspace{1mm}
	\caption{Ablations on: group attention (GA) module and the influence of the different input frames in our hierarchical information aggregation way.}
    \vspace{-5mm}
	\label{frames}
\end{table}

~~\\
\textbf{Fast Spatial Alignment.} To investigate the effectiveness and efficiency of the proposed fast spatial alignment, we equip the proposed TGA model with three different pre-alignment strategies: TGA without alignment, TGA with PyFlow~\cite{pathak2017learning}, and TGA with FSA. The evaluation is conducted on Vimeo-90K-T where there is various motion in the video clips.
Tab.~\ref{pre-align} shows the performance of TGA with PyFlow is significantly inferior than the TGA model without any pre-alignment. It implies that imperfect optical flow estimation leads to inaccurate motion compensation such as distortion on the regions with large motion (see the green box in Fig.~\ref{flow_homo}), which confuses the model during training and hurts the final video super-resolution performance. In contrast, the proposed FSA boosts the performance of the TGA model from 37.32dB to 37.59dB. This demonstrates that the proposed FSA, which although does not perfectly align frames, is capable of reducing appearance differences among frames in a proper way. We also compute time cost of this module on Vimeo-90K-T dataset and present it in Tab.~\ref{pre-align}. Our FSA method is much more efficient than the PyFlow method. Note that since every sequence in Vimeo-90K-T only contains 7 frames, the advantage of FSA in reducing redundant computation is not fully exployed. Both PyFlow and our FSA are run on CPU, and FSA could be further accelerated with optimized GPU implementation.\\ 
\begin{table}[t]
	\centering
	\scalebox{0.8}{	
		\begin{tabular}{lccc}
			
			\toprule
			Pre-alignment & w/o & w/ PyFlow~\cite{pathak2017learning}  & w/ FSA
			\\
			\midrule
			PSNR/SSIM & 37.32/0.9482 &35.14/0.9222   &\textbf{37.59}/\textbf{0.9516}
			\\
			Time (CPU+GPU) & 0+70.8ms &760.2+70.8ms & 18.6+70.8ms
			\\
			\bottomrule
		\end{tabular}
	}
     \vspace{1mm}
	\caption{Ablation on: the effectiveness and efficiency of the fast spatial alignment module. The elapsed time are calculated on processing a seven frame sequence with LR size of 112$\times$64.}
  \vspace{-1mm}
	\label{pre-align}
\end{table}

%% file: 5-conclusion.tex
\section{Conclusion}
\label{conclusion}
In this work, we proposed a novel deep neural network which hierarchically integrates temporal information in an implicit manner. To effectively leverage complementary information across frames, the input sequence is reorganized into several groups of subsequences with different frame rates. The grouping allows to extract spatio-temporal information in a hierarchical manner, which is followed by an intra-group fusion module and inter-group fusion module. The intra-group fusion module extracts features within each group, while the inter-group fusion module  borrows complementary information adaptively from different groups. Furthermore, an fast spatial alignment is proposed to deal with videos in case of large motion. The proposed method is able to reconstruct high-quality HR frames and also maintain the temporal consistency. Extensive experiments on several benchmark datasets demonstrate the effectiveness of the proposed method. 
